\documentclass[10pt,twocolumn,letterpaper]{article}

\usepackage{wacv}
\usepackage{times}
\usepackage{epsfig}
\usepackage{graphicx}
\usepackage{amsmath}
\usepackage{amssymb}
\usepackage{epstopdf}
\usepackage{bm}
\usepackage{url}
\usepackage{multirow}
\usepackage{subcaption}
\usepackage{cleveref}[2012/02/15]
\usepackage{fancyhdr}

\newcommand*{\argmax}{\operatornamewithlimits{argmax}\limits}

\crefformat{footnote}{#2\footnotemark[#1]#3}

\wacvfinalcopy 

\def\wacvPaperID{301} 

\ifwacvfinal\fi
\fancypagestyle{firststyle}
{
   
   \fancyhf{}
   \fancyfoot[C]{\raggedright \footnotesize
         \copyright  2015 IEEE. Personal use of this material is permitted. Permission from IEEE must be obtained for all other uses, in any current or future media, including reprinting/republishing this material for advertising or promotional purposes, creating new
collective works, for resale or redistribution to servers or lists, or reuse of any copyrighted component of this work in other works.  DOI: 10.1109/WACV.2015.60}
}

\begin{document}

\title{Improving Vision-based Self-positioning in Intelligent Transportation Systems via Integrated Lane and Vehicle Detection}

\author{Parag S. Chandakkar, Yilin Wang and Baoxin Li, Senior Member, IEEE \\
School of Computing, Informatics and Decision Systems Engineering 
\\Arizons State University, Tempe. \\
{\tt\small \{pchandak,ywang370,baoxin.li\}@asu.edu}
}

\maketitle
\ifwacvfinal\thispagestyle{empty}\fi
\thispagestyle{firststyle}

\begin{abstract}
   Traffic congestion is a widespread problem. Dynamic traffic routing systems and congestion pricing are getting importance in recent research. Lane prediction and vehicle density estimation is an important component of such systems. We introduce a novel problem of vehicle self-positioning which involves predicting the number of lanes on the road and vehicle's position in those lanes using videos captured by a dashboard camera. We propose an integrated closed-loop approach where we use the presence of vehicles to aid the task of self-positioning and vice-versa. To incorporate multiple factors and high-level semantic knowledge into the solution, we formulate this problem as a Bayesian framework. In the framework, the number of lanes, the vehicle's position in those lanes and the presence of other vehicles are considered as parameters. We also propose a bounding box selection scheme to reduce the number of false detections and increase the computational efficiency. We show that the number of box proposals decreases by a factor of 6 using the selection approach. It also results in large reduction in the number of false detections. The entire approach is tested on real-world videos and is found to give acceptable results.
\end{abstract}

\section{Introduction}

The United States has $786$ motor vehicles per $1000$ people, which ranks as third highest in the world \footnote{Statistics taken from \url{http://data.worldbank.org/}}. It has an estimated total of $253.1$ million registered vehicles as of 2011 according to Bureau of Transportation Statistics \footnote{Statistics taken from \url{http://www.rita.dot.gov/bts/}}. The vehicle ownership in the United States has been on constant rise with some occasional fluctuations. Traffic congestion has always been a severe problem. There are various measures developed to quantify the problem of traffic congestion, \eg congestion cost and yearly hours of delay per commuter, freeway travel time index (FTTI) etc. Large urban areas with more than $3$ million population suffer an average of $52$ hours of delay per year per auto commuter. Each commuter also has to bear the congestion cost of $\$1128$ per year. Since freeway travel is a large part of our daily commute, experts have developed measures such as FTTI. The average value of FTTI in 2011 for the freeways in large urban cities was $1.31$, which implies the freeway travel duration was increased by factor of $1.31$ per auto commuter \footnote{Congestion statistics taken from \url{http://mobility.tamu.edu/ums/national-congestion-tables/}}. The overall congestion cost and delay for $498$ metropolitan areas in the United States was $\$121$ billion and $5.5$ billion hours \cref{NationCongestion}. The problem of traffic congestion is also slowly spreading to small cities as well as rural areas.

To remedy this situation, Federal Highway Administration (FHWA) is trying to implement various policies such as dynamic traffic signal timings and varying tolls and pricing for roads with different levels of activity \footnote{\label{NationCongestion}Nationwide congestion information taken from \url{http://www.fhwa.dot.gov/congestion/}}. To apply these strategies, it is essential to know the state of the traffic on the freeway at any given instant of time. Currently, stationary loop detectors carry out the task of estimating the traffic flow at certain checkpoints with certain accuracy. However, they have reliability issues and they cannot estimate the flow of traffic on a finer resolution level, \eg lane-level traffic flow. Deploying loop detectors is expensive too \cite{herrera2007traffic}. In recent years, there has been a tendency to rely on smart ubiquitous devices such as mobile sensors \cite{herrera2007traffic} or GPS data. Though these techniques increase the overall accuracy, they cannot provide enough resolution of the traffic density.

Intelligent transportation system (ITS) research analyses the traffic flow and provides necessary feedback, mainly through varying congestion pricing and toll. The envisioned future for ITS systems is that they should be able to collect the information on lane-level and price lanes accordingly for platoons of cars entering and exiting the freeways at a given time \cite{smarTek}. This requires estimation of the number of lanes on the road and the position of the vehicle within those lanes. We call the process of determining these two parameters as \textit{vehicle self-positioning}. To calculate the (approximate) density of the vehicles, a detection module also needs to be in place. Hardware-dependent systems such as LIDAR and those using GPS are present in literature, but they are costly and are not usually preferred \cite{lidar,lidar2}.

Video-based lane and vehicle detection is popular and well-researched \cite{survey,vehicleSurvey}. Though even simple solutions such as Hough transform can extract the lane markers, inferring the road structure from them has proven difficult. The stand-alone problem of self-positioning was studied in \cite{selfPosOurs} and initial findings were reported. In this paper, we extend the problem so as to perform self-positioning and vehicle detection in a closed-loop system such that each process aids the other one and in-turn increases efficiency of the entire system. The primary purpose of this system is not to extract lane markers or vehicle detection, though we consider them as sub-problems, but to perform accurate self-positioning with the aid of vehicles present on the road. We propose a Bayesian model which takes three parameters: number of lanes on the road, the lane in which vehicle is being driven and the presence of a vehicle. To construct the likelihoods of this model, we use a bottom-up approach which uses guided-filter, lane-model generation and a vehicle detection module. To allow further exploration of this problem as well as to improve and verify current techniques, we have made the database publicly available.

The organization of rest of the paper is as follows. Section 2 discusses recent relevant literature on lane and vehicle detection and self-positioning. Section 3 describes the proposed approach in detail. Section 4 explains the experimental setup and results. Section 5 gives concluding remarks and outlines the future scope for the current system.


\section{Related Work}
Lane and vehicle detection are sub-tasks within our current problem formulation. Very few attempts have been made in the past to perform self-positioning, however there are many algorithms reported in the literature which handle the problem of lane and vehicle detection separately. Both software and hardware-dependent approaches have been developed for the said tasks. Here we focus only on vision-based approaches.

Lane detection includes multiple stages, namely, image pre-processing, feature extraction, model-fitting for further verification and tracking to maintain spatio-temporal consistency. One of the most popular pre-processing methods is inverse perspective mapping (IPM) which maps an image to a bird's eye-view. This makes the lane markers appear straight and parallel to each other. Also, other fine details in the image get suppressed. Next step is usually simple morphological operation followed by parametric line fitting to extract the lane markers \cite{line2,line3,line4}. Kalman filter is also used for tracking to minimize the effect of false detections \cite{recBayesian}. In another tracking-based method, the lane model generated from previous frame was enforced onto the new frame and only newly appearing lanes were considered \cite{tracking1}. Since roads are not straight, some methods have tried to fit splines or higher-order curves to get an accurate representation of lane markers \cite{line6,gradient2}. However, fitting higher-order curves or B-snakes may be computationally inefficient. Another interesting method which can be used to detect more than one lane (host-lane) is the hierarchical bi-partite graph-based road modeling \cite{hbpg}. This approach outputs multiple lanes and tries to understand the road structure. This approach also assigns a confidence measure for each lane. However, this method does not attempt to position the vehicle in the lanes.

A similar problem to self-positioning has been handled in \cite{kuhnl2013visual} using spatial RAY features. It tries to predict the position of the vehicle on the road from an input video stream. However, they predict at most 3 lanes at a time either on the left or right hand side, whereas we have upto 6 lanes in our database with upto 3 lanes being on each side. In addition, there experiments are spread over only 2 days with similar road conditions. Our experiments are more rigorous, spreading over 5 days and the data includes 6 different road conditions. They have low traffic density whereas we have moderate-to-high traffic density. Thus direct comparison between these two approaches may not be possible.

Due to recent progress in computer vision research and computational abilities, patch-learning based approaches classify image patches depending on the presence of lane-markers. Gradients and steerable Gaussian filters act as good features for such patch-based methods \cite{steerable}. Attempts are also made to learn the road structure by collecting knowledge about the type and structure of lane-markers \cite{learning1}.

Vehicle detection is a subset of a widely-studied problem of object detection. There are a plethora of approaches for general object detection, \eg \cite{lsvm-pami}, or approaches dedicated to vehicle detection which use optical-flow and hidden-markov-model based classification to interpret motion-based clues \cite{opticalFlowHMMCar}. HOG and HAAR features have also been used with different classifiers and learning frameworks such as adaboost, SVM and active learning \cite{activeLearningCar,yuan2011learning}. Though there are innumerable methods developed for general object detection and specifically for vehicle detection, listing them all is beyond the scope of this paper. A good review of vehicle detection methods can be found in \cite{vehicleSurvey}. In spite of continuous efforts, there are no reliable vision-based solutions to predicting number of lanes and vehicle self-positioning given a front-facing view.

\section{Proposed approach}

In this section, we propose an integrated approach to solve the problem of vehicle self-positioning. Apart from external factors such as bad weather and conditions of lane markings, lane-occlusion by passing vehicles is one of the biggest hurdles to reliable vehicle self-positioning. Our approach works in a closed loop. We utilize information from the positions of other vehicles to improve our self-positioning. The self-positioning information gives us information about the road structure, which we use to generate the detection proposals for vehicles. We also employ temporal smoothing of results across frames to counter the effect of unexpected events, \eg a car stopped on the shoulder or faint/invisible lane markings.

We incorporate this problem in a Bayesian framework to add the additional factors such as presence of vehicles, vehicle dynamics, scene semantics etc. Assume we have labeled data $D=\begin{bmatrix}
						 F_1 & \Theta_1 \\
 						 F_2 & \Theta_2 \\
  						 \vdots & \vdots \\
						 F_n & \Theta_n \\
						 \end{bmatrix}$ where $F_i$ is the $i^{th}$ video clip and $\Theta_i$ is a $2-D$ label vector $[\theta_1, \theta_2]$ where $\theta_1 \in [1, n_{lanes}]$ is the number of lanes present in $F_i$. The other parameter $\theta_2 \in [1, \theta_1]$ is the lane in which the car is currently driving (assuming the leftmost lane is the first lane). The process of determining the above label vector for each frame is called self-positioning. Each video clip $F_i$ contains $m$ frames - $[f_i^1,f_i^2,\hdots,f_i^m]$. We include the vehicle presence in the Bayesian framework in order to aid self-positioning as follows. The joint probability density for any frame $f_k^j$ such that the video clip $F_k$ belongs to the test data is $P(f_k^j,\Theta \vert D,f_k^{j-1},\hdots,f_k^{j-p},V_i)$ where $[f_k^{j-1},f_k^{j-2},\hdots,f_k^{j-p}]]$ is a set of $p$ frames occurring before $f_k^j$ and $V_i$ denotes the vehicle presence in lane $i$. We need to predict the label vector $\hat{\Theta}$ for current frame $f_k^j$ given the set of previous $p$ frames, the training data D and vehicle presence in a lane $i$ can be written as, $P(\Theta \vert D,f_k^{j-1},\hdots,f_k^{j-p},V_i)$. By Bayes' rule it can be further decomposed into,					
\begin{multline}
P(\Theta \vert D,f_k^{j-1},\hdots,f_k^{j-p},V_i) \propto \\ P(f_k^{j-1},\hdots,f_k^{j-p} \vert \Theta,D,V_i) P(V_i \vert \Theta,D).
\end{multline}

We assume a uniform prior on presence of vehicles. Also, we introduce a temporal smoothing kernel which considers the results of previous frames. This helps to minimize the effect of sudden external factors as explained before. The above maximum-a-posteriori (MAP) problem can be formulated into a maximum-likelihood estimation (MLE) problem as,
\begin{align*}
  \hspace{2em}&\hspace{-233.5pt}\hat{\Theta} \hspace{2pt} = \hspace{2pt}\argmax_{\Theta} \hspace{2pt} P(f_k^j,\hdots,f_k^{j-p} \vert \Theta,D,V_i)\\
  \hspace{8pt} \approx \hspace{2pt} \argmax_{\Theta} \hspace{2pt} \psi^T \hspace{2pt}[P(f_k^j \vert \Theta,D,V_i) \hdots P(f_k^{j-p} \vert \Theta,D,V_i)]
\end{align*}
\noindent
The temporal smoothing kernel takes the form of a slowly increasing exponential function as shown in the following equation.
\begin{multline}
\psi(i)=\text{initial value} \times (1+\text{rate-of-increase})^i \\ \forall \hspace{2pt} i \in [1,p+1]
\end{multline}
where $i$ is the frame index and $p$ is the number of frames we have buffered. The values are normalized so that the $(p+1)^{th}$ frame (which is also the current frame) gets a unit weight. Now, expanding only the first term from the previous step (for readability purposes), we get,
\begin{align*}
\hspace{2em}&\hspace{-5em}\hat{\Theta} \hspace{2pt} = \hspace{2pt} \argmax_{\Theta} \psi^T \hspace{2pt} P(f_k^j \vert \Theta,D,V_i) \hspace{3.5pt} P(\Theta \vert V_i,D) \\
\hspace{2em}&\hspace{-4.05em} = \argmax_{\Theta} \psi^T \hspace{2pt} P(f_k^j \vert \Theta,D,V_i) \hspace{2pt} P(\theta_1,\theta_2 \vert V_i,D)
\end{align*}
\noindent
The determination of our current lane does not depend on the vehicle presence in a lane since we do not incorporate view-point information of a detected object. Thus we take $P(\theta_2 \vert V_i,D)$ as constant and remove it from the equation.

\begin{align*}
\hspace{2em}&\hspace{-4.65em}\hat{\Theta} \hspace{2pt} = \argmax_{\Theta} \psi^T \hspace{2pt} P(f_k^j \vert \Theta,D,V_i) \hspace{2pt} P(\theta_1 \vert V_i,\theta_2,D)
\end{align*}

\noindent
The term $P(f_k^j \vert \Theta,D,V_i)$ involves calculating features of the current frame given the label vector and the vehicle detection result. However, the detection result only affects the label vector, and in particular $\theta_1$. The feature extraction process and the vehicle detection process run parallely and so they are independent of each other. By definition of $\theta_1$ and $\theta_2$, we see that though $\theta_2$ is weakly conditioned on $\theta_1$ as $\theta_2 \in [1,\theta_1]$, vice versa does not hold (i.e. the current lane in which we are driving does not determine the number of lanes on the road). Therefore we remove the condition of $\theta_2$ from the last term. We also re-introduce the previously skipped terms containing frames $[f_k^{j-1},\hdots,f_k^{j-p}]$ in the following equation.
\begin{multline} \label{eq:genFormulatn}
\hat{\Theta} = \argmax_{\Theta} \psi^T \hspace{2pt} [P(f_k^j \vert \Theta,D) \hspace{3.5pt} P(\theta_1 \vert V_i,D) \cdot \hdots \cdot \\ P(f_k^{j-p} \vert \Theta,D) \hspace{3.5pt} P(\theta_1 \vert V_i,D)].
\end{multline}

Equation \ref{eq:genFormulatn} shows a general formulation to obtain the correct value of the label vector given a set of buffered frames and the vehicle detection results. The structured Bayesian formulation allows easy modification of the formulation if we were to add more information to the model in the future. In the next paragraphs, we show how to calculate the two quantities - $P(f_k^j \vert \Theta,D)$ and $P(\theta_1 \vert V_i,D)$.

\subsection{Frame likelihood computation}

\begin{figure*}[!t]

\centering
\includegraphics[scale=0.628]{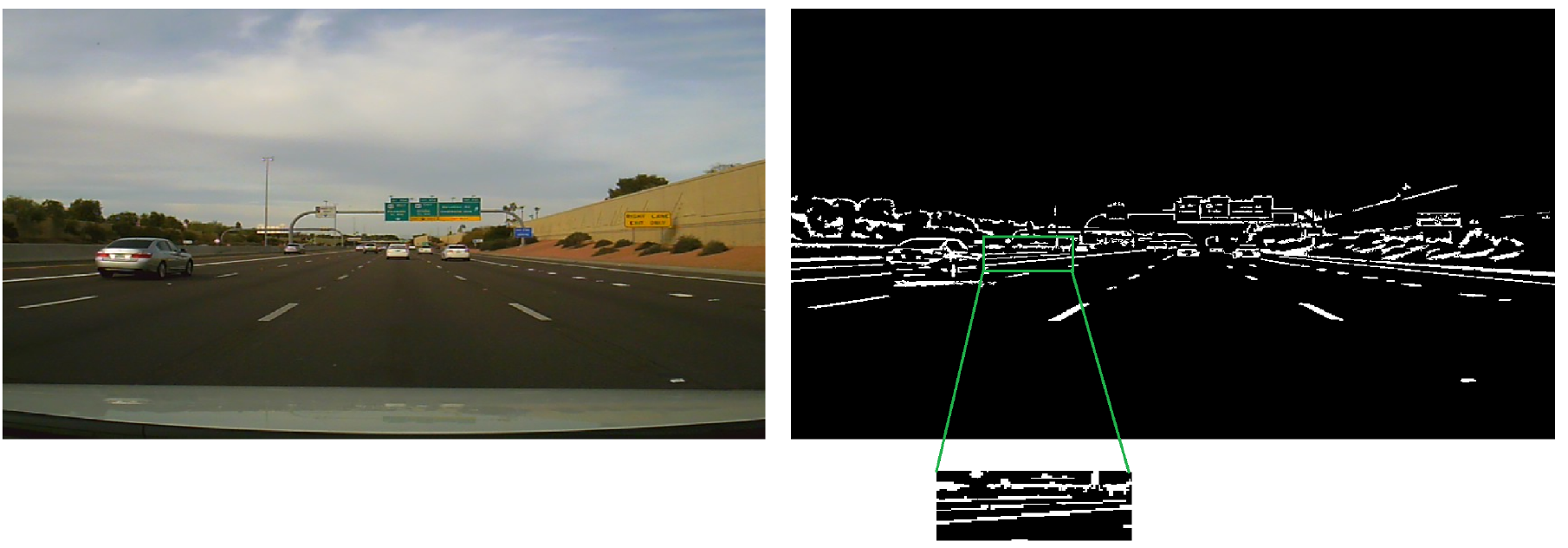}
\caption{Captured frame and pre-processing using guided filter.}
\label{fig:guided}
\end{figure*}

The frame likelihood computation involves extracting features from the current frame and then calculating an initial estimate of the label vector, which is refined later by using vehicle detection results. It includes the following stages:

\begin{enumerate}
\item Image pre-processing.
\vspace{-4pt}
\item Lane model generation.
\vspace{-4pt}
\item Feature extraction.
\vspace{-4pt}
\item Initial estimation of frame likelihood.
\end{enumerate}

\begin{figure}[!b]
\centering
\includegraphics[width=0.48\textwidth,height=130pt]{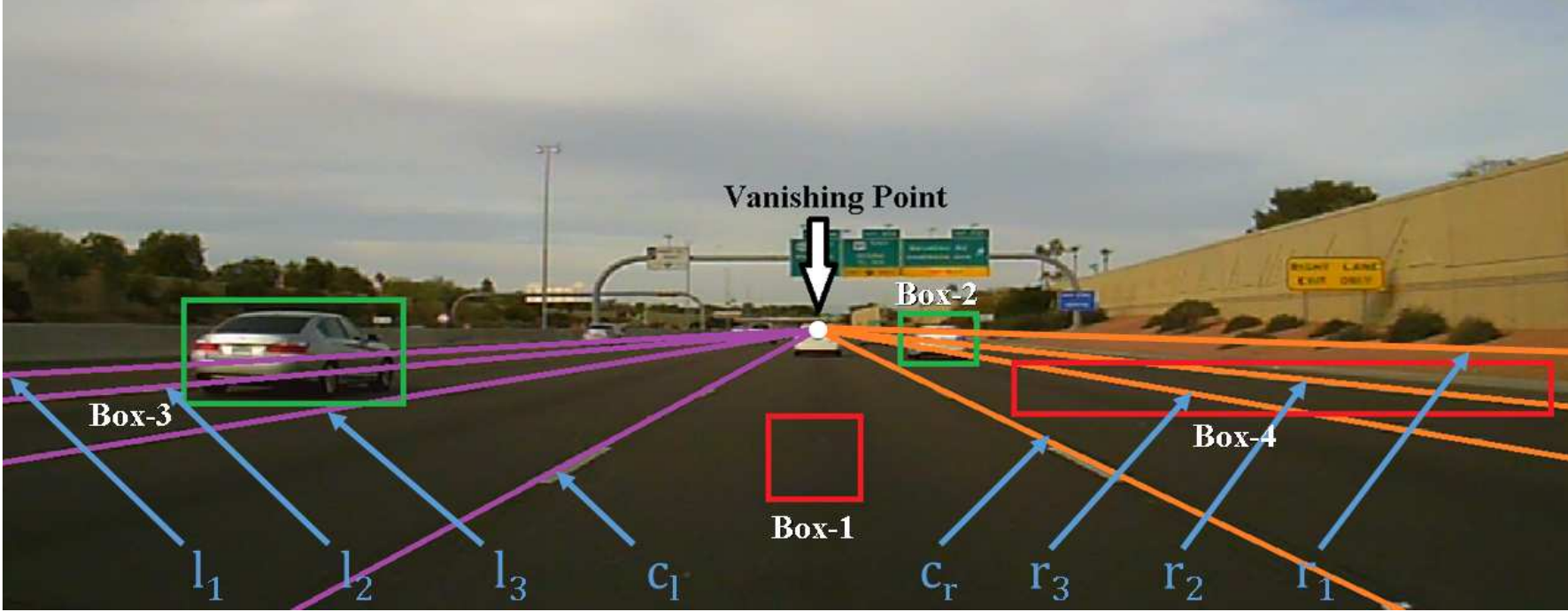}
\caption{Lane width modeling}
\label{fig:laneWidthModel}
\end{figure}

\noindent
\textbf{Image pre-processing:} Our aim is to detect all the lanes present in the frame in order to compute the label vector $\Theta$. Pre-processing an image removes the unnecessary details in it and keeps all the lanes. We choose Guided filter for this task. It is a edge-preserving smoothing filter which works in linear time \cite{guided}. It takes a pair of images as input. One of them acts as a reference image and ``guides" the filtering process of the other image. When both images are same, the filter performs edge-preserving smoothing. The filtering operation is defined as:

\begin{equation}
GF(I)=\bar{\alpha} I + \bar{\beta},
\end{equation}

\noindent
where $I$ is a gray-scale video frame.

$\bar{\alpha}$ and $\bar{\beta}$ are obtained through block-wise averaging using $\alpha=\dfrac{\sigma^2}{\sigma^2+\epsilon}$ and $\beta=(1-\alpha) \mu$, where $\mu$ is the mean, $\sigma$ is the standard deviation of a block in the image and $\epsilon$ is a small constant. For a flat patch in an image, $\sigma=0 \implies \alpha=0$ and $\beta=\mu$. Thus each pixel in that flat region is averaged with its neighbors. Similarly it can be proved that if $\sigma \gg \epsilon$, sharp edges are preserved. However, by choosing $\epsilon$ appropriately, we can force the guided filter to consider almost all the pixels in an image belonging to a flat patch. We choose $\epsilon$ high enough such that lanes are considered belonging to a flat patch. The lane markings are always surrounded by the road pixels which have much lower value and therefore, the value of lane pixels decreases by a large amount as compared to anywhere else in the image. Now a post-processing operation such as over-subtracting followed by saturation returns an image similar to the one shown in Fig. \ref{fig:guided}. The post-processing operation we use is as follows:

\begin{equation}
BW(I)=[(I-\delta \cdot GF(I))*255] \geq 0,
\end{equation}

\noindent
where $\delta$ is a constant just greater than 1 (here, $\delta=1.06$). Since value of lane marking pixels has reduced by a large extent, they get preserved in the post-processed image whereas many other regions disappear. Advantage of guided filter is that it can also work in bad weather conditions such as rain, low-sunlight or in the night. It can also detect extreme lanes which are very thin as shown in Fig. \ref{fig:guided}.

\vspace{3.5pt}
\noindent
\textbf{Lane model generation:} The pre-processed image preserves all the lanes and rejects a lot of unwanted regions. Yet, there are many spurious objects which may prevent us from performing reliable detection of all the lanes on the road. Additionally, as mentioned before, our aim is to perform self-positioning only on the freeways which have a defined road structure such as constant-width lanes. Taking advantage of these constraints, we define a lane model in which there are a maximum of seven lanes, three being on each side of the center lane. Due to fixed-width of the lanes, they can be represented as a function of y-coordinate of the center lane markers and the camera parameters. We first detect the center lane markers in an image using simple thresholding techniques and other heuristics depending on their possible position, shape and color. We then perform a linear fit for the left and right center lane marker denoted respectively by $c_l$ and $c_r$. Then the remaining three lane markers on each side can be obtained as:

\begin{equation} \label{eq:leftLane}
l_i=d_{l_i}(c_l)=m_i^l \hspace{1pt} y_{c_l} + k_i^l, \hspace{5pt} \forall i \in [1,2,3]
\end{equation}
\noindent
and,
\begin{equation} \label{eq:rightLane}
r_i=d_{r_i}(c_r)=m_i^r \hspace{1pt} y_{c_r} + k_i^r, \hspace{5pt} \forall i \in [1,2,3]
\end{equation}

\begin{figure}[!b]
\centering
\includegraphics[width=0.35\textwidth,height=85pt]{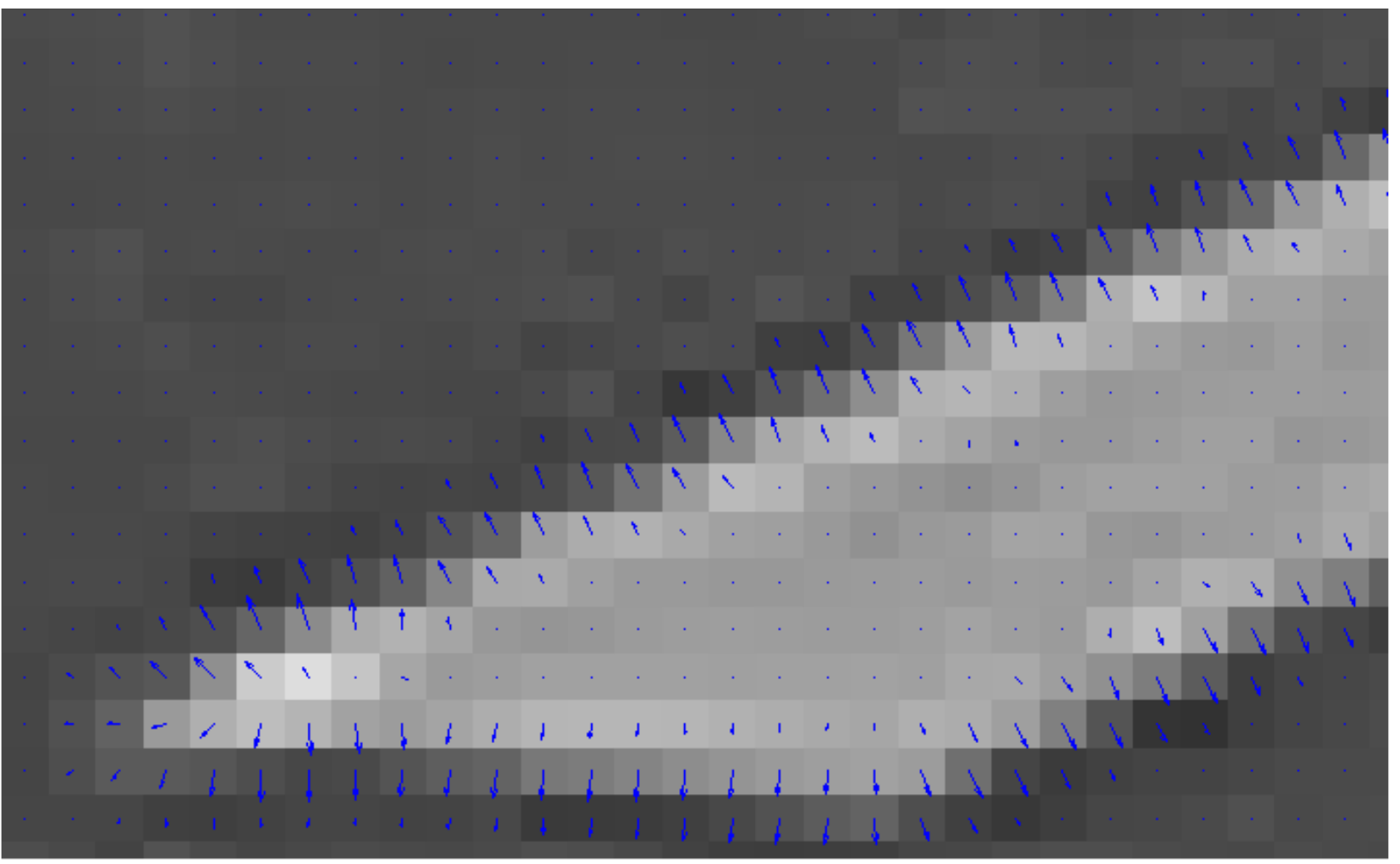}
\caption{Obtaining road pixels from lane pixels}
\label{fig:roadPixelDetection}
\end{figure}

Above two equations represent the other lanes in the form of offsets from the center lane - $d_{l_i}(c_l)$ and $d_{r_i}(c_r)$. These offsets are in turn represented as a linear function of the y-coordinate the center lane markers - $y_{c_l}$ and $y_{c_r}$. $\{m_i^l,m_i^r,k_i^l,k_i^r\}$ are the slope and intercept parameters of $c_l$ and $c_r$ respectively. Once these offsets have been calculated, then obtaining $(x,y)$ coordinates of lane-markers is a straight-forward task. We assume camera parameters are fixed, but they can also be included in equation \ref{eq:leftLane} and \ref{eq:rightLane}. The generated lane model is shown in Fig. \ref{fig:laneWidthModel}.

\vspace{4pt}
\noindent
\textbf{Feature extraction:} Once we have found the probable lane regions as shown in Fig. \ref{fig:laneWidthModel}, we find lane pixels by simple thresholding. By following the gradient directions at those lane pixels, we can get the road pixels as shown in Fig. \ref{fig:roadPixelDetection}. We form a 40-dimensional feature vector consisting of:

\begin{enumerate}
\item Mean and variance of lane and road pixels (4-D).
\vspace{-4pt}
\item 36-bin histogram of gradients at lane pixels as shown.

\end{enumerate}

The lane markers will have low mean and variance for road pixels and the majority of the gradients at lane pixels lie in a specific range of angles (shown in Fig. \ref{fig:roadPixelDetection}). We consider at most 7 lanes (or 8 lane markers) at a time, upto 3 lanes being on either side. Presence of middle lane is assumed. Thus our feature vector is $40*6=240$ dimensional.

\vspace{4pt}
\noindent
\textbf{Estimation of $\bm{P(f \vert \Theta,D)}$: }Though there are many methods to implement the general formulation presented in equation \ref{eq:genFormulatn}, we choose to implement it using a linear SVM. It is trained using the features extracted from $D$. For a video frame in the test data, we extract its features and then apply the linear SVM. The likelihood estimate $\mathcal{L}(\Theta \vert f,D)$ is the probability estimate of the linear SVM for each $\Theta$. We repeat the same procedure with random forest too.

\subsection{Refinement of frame likelihood}
We use the presence of vehicles in the adjacent lanes to our advantage. The term $P(\theta_1 \vert V_i,D)$ shows that the vehicle presence in lane $i$ affects the probability of the number of total lanes on the road. Assume an initial estimate of the label vector $\hat{\Theta}_{init}=[\hat{\theta_1}_{init}, \hat{\theta_2}_{init}]$. Here, $\hat{\theta_1}_{init}$ and $\hat{\theta_2}_{init}$ are the initial estimates for the number of lanes and the host lane respectively. From $\hat{\Theta}_{init}$ we can obtain indices of lanes which are present, \eg if $\hat{\Theta}_{init}=[5, 2]$ then the lane indices are $L_{ind}=\{l_3,c_l,c_r,r_3,r_2,r_1\}$ or $\{3,4,5,6,7,8\}$ (refer Fig. \ref{fig:laneWidthModel}). Depending on the vehicle presence in lane $i$, we refine $\hat{\Theta}_{init}$ in the following manner:

\begin{equation}
\hat{\theta_1}_{new}=\begin{cases}
	\hat{\theta_1}_{init} \hspace{20pt} \text{if} \hspace{4pt} min(L_{ind}) \leq \hat{\theta_1} \leq max(L_{ind}) \\
	\hat{\theta_1}_{init}+(min(L_{ind})-i)  \hspace{29.5pt} \text{if} \hspace{4pt} 1 \leq i<4 \\
	\hat{\theta_1}_{init}+(i-max(L_{ind}))  \hspace{28pt} \text{if} \hspace{4pt} 5 < i \leq 8
	\end{cases}
\end{equation}
	
This requires a reliable vehicle detector with high quality object proposals to start with. In the next section we outline a method to generate high-quality object proposals by using the lane structure shown in Fig. \ref{fig:laneWidthModel}.

\begin{figure*}[!t]
        \centering
        \begin{subfigure}[!t]{0.3\textwidth}
                \includegraphics[width=\textwidth]{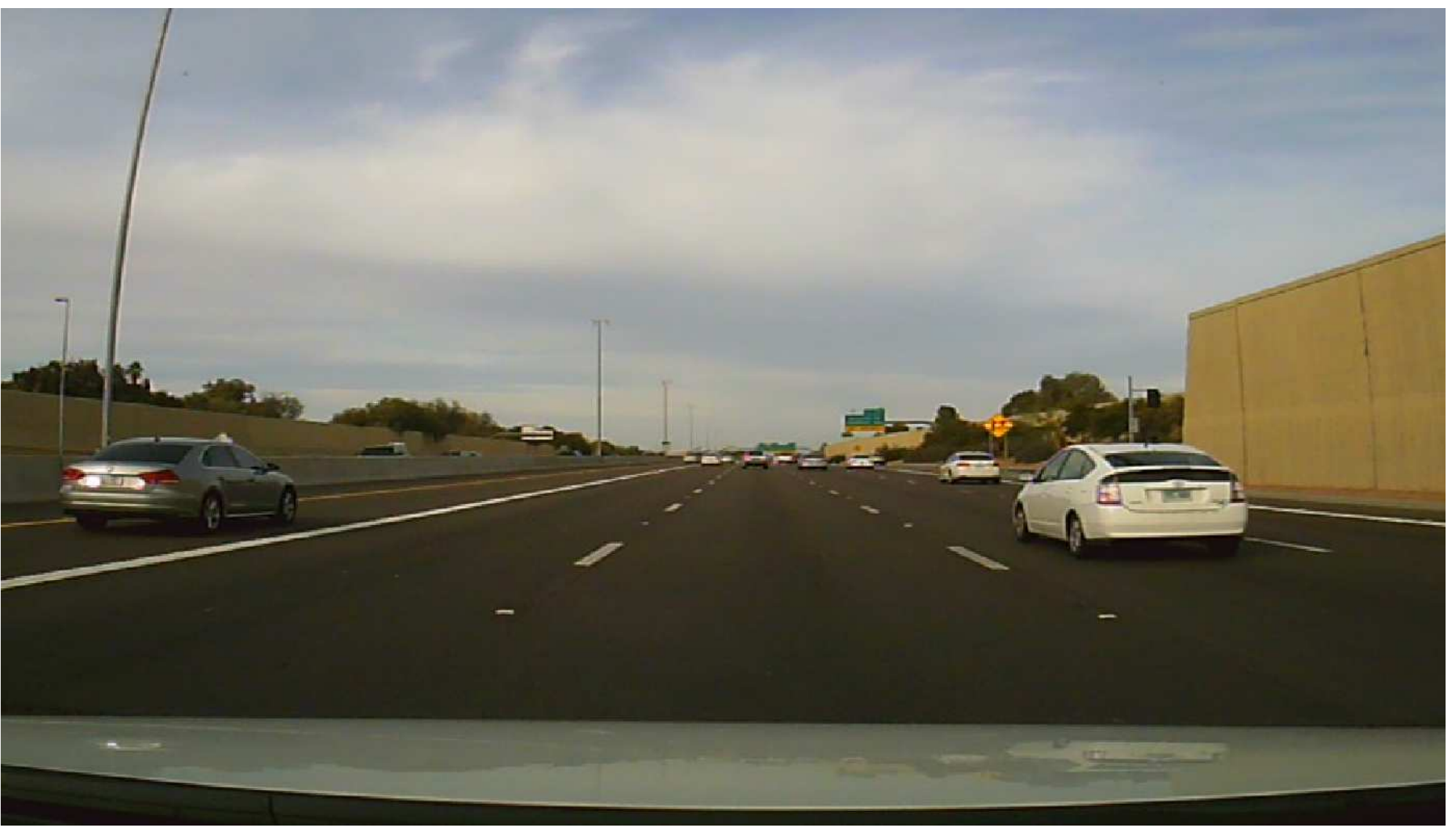}
        \end{subfigure}%
        \hspace{1pt}
        ~ 
        \begin{subfigure}[!t]{0.3\textwidth}
                \includegraphics[width=\textwidth]{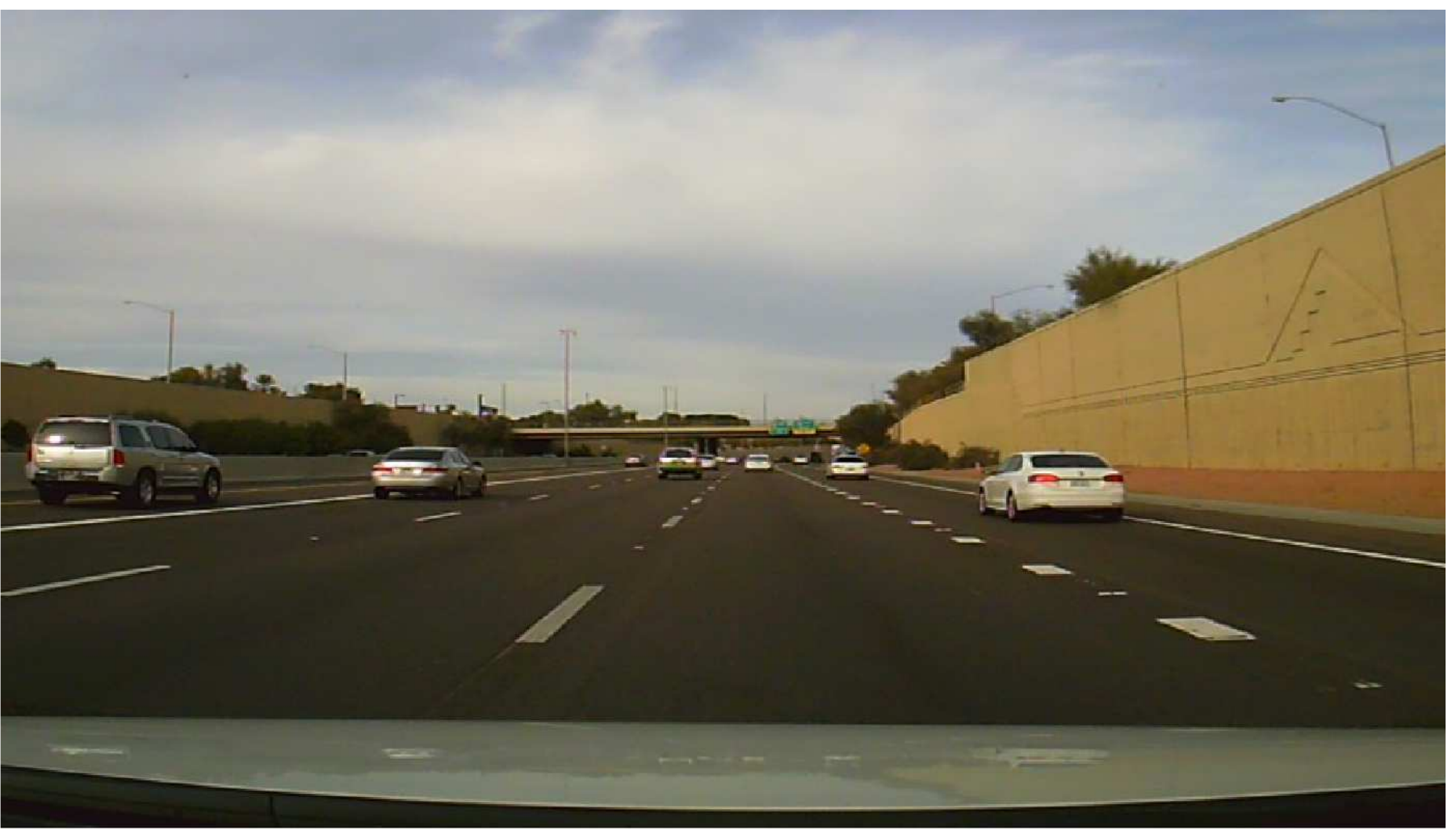}
        \end{subfigure}
        ~ 
        \begin{subfigure}[!t]{0.3\textwidth}
                \includegraphics[width=\textwidth]{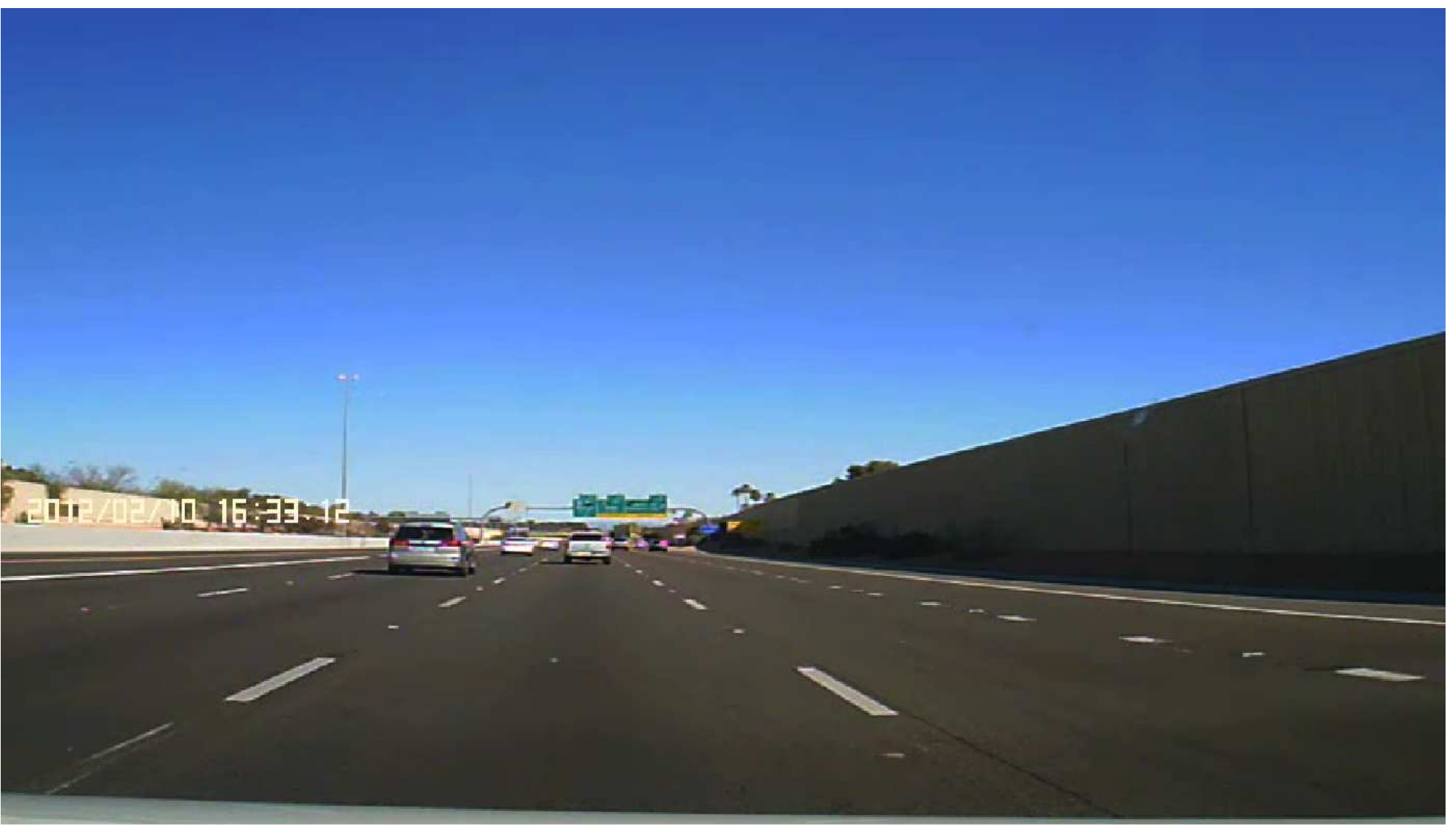}
        \end{subfigure}

		\vspace{2pt}

        \begin{subfigure}[!t]{0.3\textwidth}
                \includegraphics[width=\textwidth]{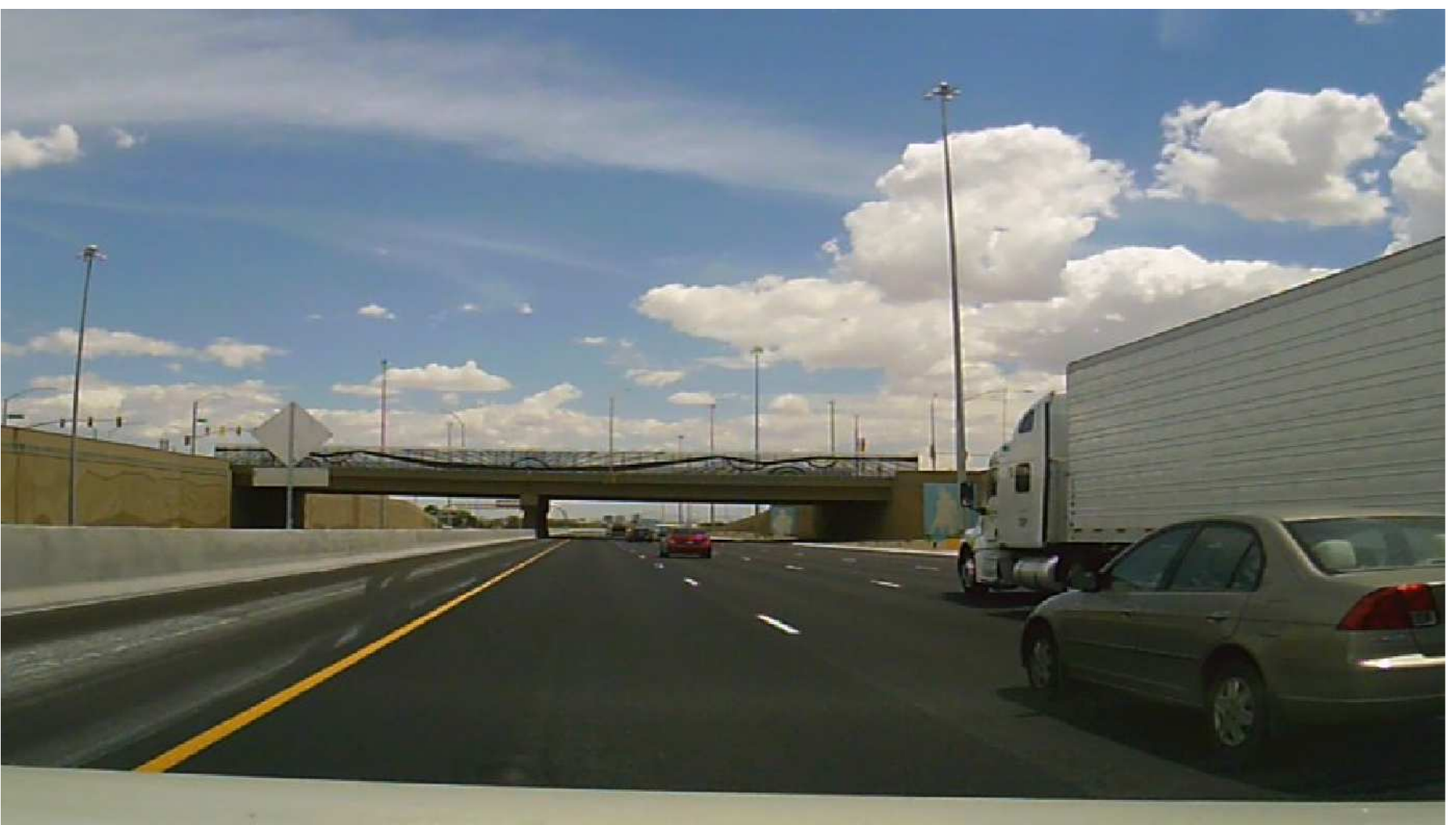}
        \end{subfigure}%
        \hspace{1pt}
        ~ 
        \begin{subfigure}[!t]{0.3\textwidth}
                \includegraphics[width=\textwidth]{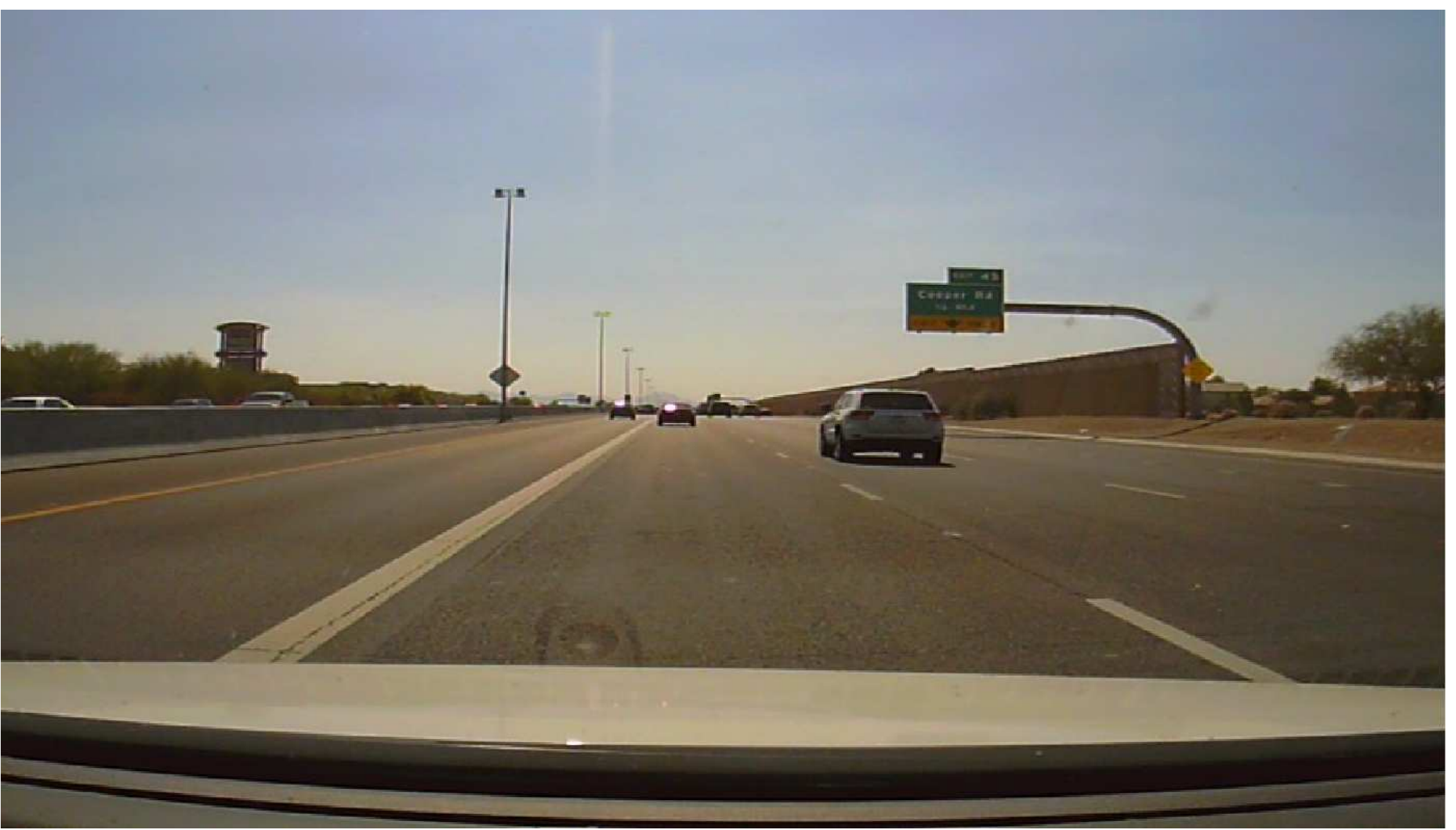}
        \end{subfigure}
        ~ 
        \begin{subfigure}[!t]{0.3\textwidth}
                \includegraphics[width=\textwidth]{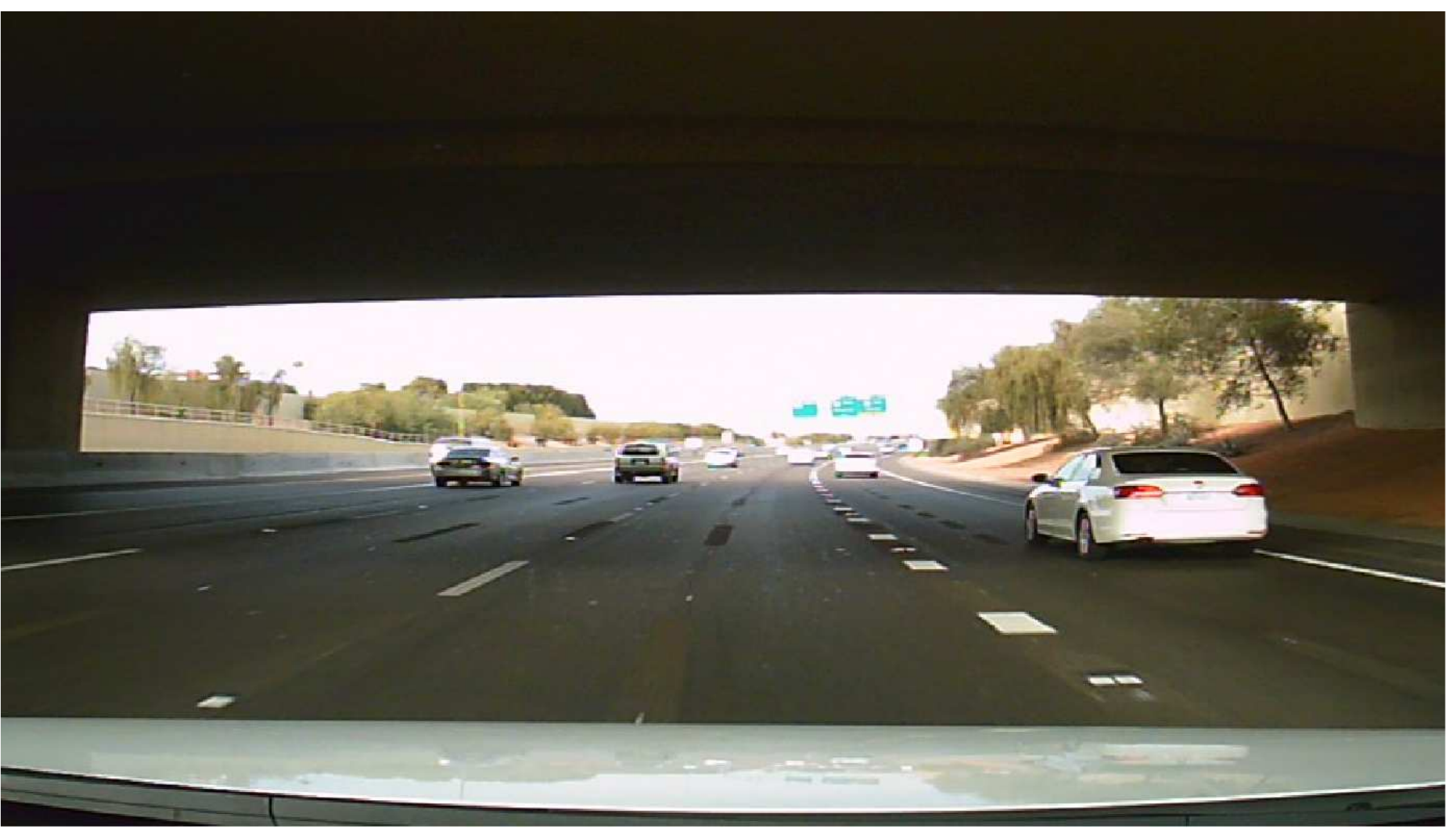}
        \end{subfigure}
        \caption{From left to right, row 1: Correct predictions for categories $[5,3],[5,4]$ and $[6,4]$. Category of the middle image is predicted correctly in spite of a vehicle occluding the view of first lane. Row 2: Wrong predictions for categories $[4,1]$, $[5,2]$ and $[5,4]$. Occlusion, bad road conditions and bad lighting conditions are the plausible causes for wrong detections.}
        \label{fig:rightWrongDetections}
\end{figure*}

\subsection{Efficient vehicle detection}
Our proposed approach works in a closed loop such that it first detects the vehicles present in the frame which helps us to perform self-positioning task. Then the feedback from the inferred lane structure is used to generate high-quality object proposals. The obtained detections from these proposals then traverse to the beginning of the loop.

We make a key observation from Fig. \ref{fig:laneWidthModel} that the span of lower edge of a bounding box enclosing a vehicle is always contained with the two corresponding lane markers. As shown in Fig. \ref{fig:laneWidthModel}, bounding boxes having a span too small (Box-1) or too large (Box-4) than their corresponding lane markers are invalid and are rejected in our proposal generation process. We now show a sample process of calculating the span between the lane markers and also focus on bounding box selection process.

Consider a bounding box represented by $[x_{min},y_{min},x_{max},y_{max}]$. To find the span between its corresponding lane markers, we first need to find their $(x,y)$ coordinates. Since any lane marker can be represented in as a function of the center lane markers, their coordinates $(x_c,y_c)$ can be found by setting $y_c=y_{max}$ and $x_c$ can be found using the linear fit. The nearest lane marker to $(x_{min},y_{max})$ is then found. The coordinates of that lane marker $(x_{near},y_{near})$ are found by setting $y_{near}=y_{max}$ and $x_{near}=x_{c_r}+r_j$. If the nearest lane marker is to the left then it is similarly obtained as, $x_{near}=x_{c_l}-l_j$ (please refer to equation \ref{eq:leftLane} and \ref{eq:rightLane}). Once the coordinates for the nearest lane marker are found, then the span between that and the next marker can be easily calculated.

The advantage of the above bounding box selection technique is that it involves only simple computations and it can be applied to exhaustively generated bounding boxes or to the proposals generated through techniques such as BING \cite{bing}. However, we observe that in our data-set, the video frames are not of high-quality and many vehicles are small. BING does not capture all the cars and thus we have used exhaustively generated bounding box along with the selection technique. The exhaustive boxes are generated by applying three naive constraints:

\begin{enumerate}
\item Aspect ratio of any box cannot be more than three and smaller than one-third.
\vspace{-4pt}
\item The boxes lie only in the lower $\frac{2}{3}^{rd}$ part of an image.
\vspace{-4pt}
\item Maximum size of boxes is pre-determined.
\end{enumerate}

\noindent
Even after applying these constraints, the selection technique reduces the no. of boxes by a factor of $6$ on an average and eliminates many false detections. We use a cascaded deformable parts model \cite{voc-release5,lsvm-pami} trained on our data which consists of $833$ vehicles and on ``car" subset of Pascal VOC 2010 data \cite{voc2010}. The negative data is randomly chosen. These detections can again be used to refine the frame likelihood.

%

\section{Experimental Setup and Results}

\begin{table*}[!t]
\centering
\caption{Self-positioning accuracy.}
\label{table:accuracy}
\vspace{4pt}
\begin{tabular}{| c | c | c | c | c |} \cline{2-3} \cline{4-5}

\multicolumn{1}{c|}{}
& \multicolumn{2}{ c| }{Without temporal smoothing}

& \multicolumn{2}{ c| }{With temporal smoothing} \\
\hline

\multicolumn{1}{|p{2.5cm}|}{\centering \hspace{8pt} Class = $[\theta_1,\theta_2]$} & \multicolumn{1}{|p{2.5cm}|}{\centering Linear SVM} & \multicolumn{1}{|p{2.5cm}|}{\centering Random forest} & \multicolumn{1}{|p{2.5cm}|}{\centering Linear SVM} & \multicolumn{1}{|p{2.5cm}|}{\centering Random forest}  \\
 \hline \hline
$[4,1]$      &    33.55\%  &   \textbf{34.94}\%  &  32.90\%  &  32.81\%   \\
$[4,2]$		 &    58.84\%  &   80.28\%  &  62.67\%  &  \textbf{89.32}\%   \\			
$[4,3]$		 &    69.74\%  &   73.50\%  &  73.33\%  &  \textbf{74.88}\%   \\			
$[4,4]$	     &    72.45\%  &   90.48\%  &  73.47\%  &  \textbf{96.94}\%   \\
$[5,2]$		 &    54.93\%  &   47.00\%  &  \textbf{63.44}\%  &  63.06\%   \\		
$[5,3]$		 &    48.18\%  &   47.94\%  &  49.91\%  &  \textbf{53.78}\%   \\			
$[5,4]$		 &    44.06\%  &   56.82\%  &  46.27\%  &  \textbf{64.27}\%   \\			
$[6,4]$		 &    95.45\%  &   94.55\%  &  \textbf{100}\%  &  96.36\%     \\
\hline
\multicolumn{1}{|p{3cm}|}{\centering Overall Accuracy}  &  54.77\% &  61.41\% &  57.49\% & \textbf{66.45}\% \\		
\hline
\end{tabular}
\end{table*}

\begin{table*}[!t]
\centering
\caption{Confusion matrix for self-positioning without temporal smoothing (using Random forest). Results of self-positioning after temporal smoothing are shown in brackets.}
\label{table:RFConfMat}
\begin{tabular}{| c || c | c | c | c | c | c | c | c |}
\hline
\multicolumn{1}{|p{1.4cm}||}{\centering Class = \\ $[\theta_1, \theta_2]$} 	& \multirow{2}{*}{\centering $[4,1]$} & \multirow{2}{*}{\centering $[4,2]$} & \multirow{2}{*}{\centering $[4,3]$} & \multirow{2}{*}{\centering $[4,4]$} & \multirow{2}{*}{\centering $[5,2]$}  &  \multirow{2}{*}{\centering $[5,3]$} & \multirow{2}{*}{\centering $[5,4]$} & \multirow{2}{*}{\centering $[6,4]$} \\
 \hline \hline
$[4,1]$		  &    378 (355)  &  300 (302)  &  47 (10)  &  5 (0)  &  348 (415)  &  2 (0)  &  2 (0)  &  0 (0) \\
$[4,2]$ 	  &    24 (3)  &  985 (1096) &  72 (32)  &  6 (10)  &  89 (86)  &  49 (0) &  0 (0) &  2 (0)\\
$[4,3]$		  &    10 (0)  &  12 (5) &  1659 (1690)  &  37 (35) &  2 (0) &  295 (328) &  239 (199) &  3 (0)\\			
$[4,4]$		  &    0 (0) &  9 (0) &  0 (9) &  266 (285)  &  0 (0) &  0 (0) &  17 (0) &  2 (0)\\
$[5,2]$		  &    19 (5) &  218 (184) &  23 (0) &  2 (2)  &  243 (326)  &  11 (0) &  1 (0) &  0 (0)\\		
$[5,3]$		  &    10 (0) &  515 (627) &  235 (98) &  3 (0) &  1 (3) &  779 (874)  &  46 (0) &  36 (23) \\		
$[5,4]$		  &    21 (0)  &  481 (506)  &  213 (141)  &  6 (0)  &  31 (0)  &  7 (0) &  1029 (1164) &  23 (0)\\			
$[6,4]$		  &    0 (0)  &  0 (0)  &  0 (0) &  0 (0) &  0 (0) &  1 (0) &  11 (8) &  208 (212) \\		

\hline
\end{tabular}
\end{table*}

\begin{figure}[!b]
\centering
\includegraphics[width=0.48\textwidth,height=140pt]{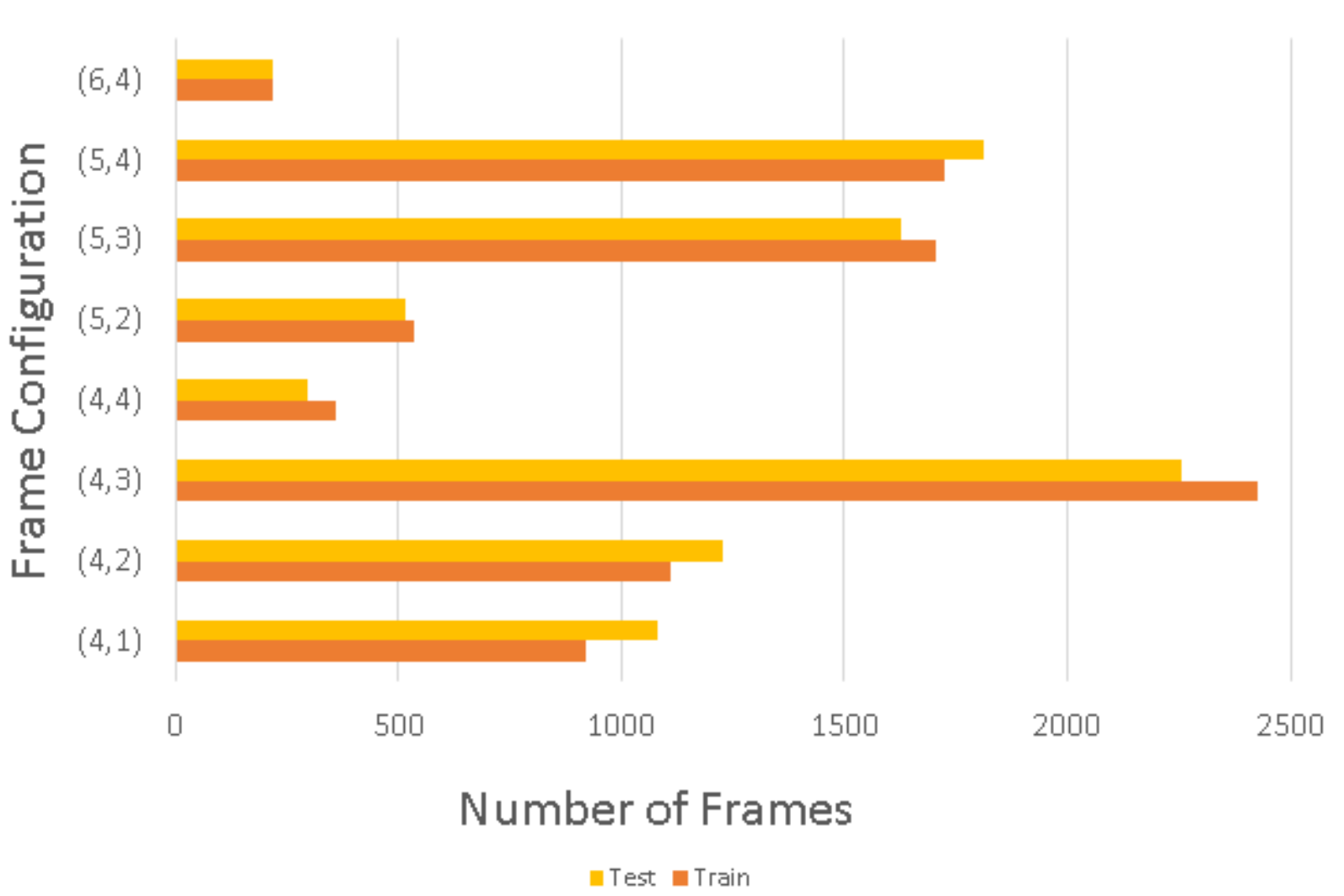}
\caption{Distribution of Video Frames in Train and Test Set}
\label{fig:trainTestDistr}
\end{figure}

We have formed the data-set using the videos captured from a typical camera mounted inside the wind-shield. Our data-set has 53 videos with a frame rate of 30 fps \footnote{The data-set is available at \url{http://www.public.asu.edu/~bli24/CodeSoftwareDatasets.html}}. The average length of each video clip is 11.8 seconds. Though $L=7$ yields $28$ categories, we consider a subset of those categories which occur frequently in real-life. The list of label configurations in our database is as follows: $\Theta = \{[4,1],[4,2],[4,3],[4,4],[5,2],[5,3],[5,4],[6,4]\}$. We assume a uniform prior for all these categories and zero prior for the rest. Our training data-set contains $27$ videos while the rest $26$ are used for testing. The number of frames used in training and testing are 9018 and 9036 respectively. The distribution of training and test data over various categories is shown in Fig. \ref{fig:trainTestDistr}. As mentioned before, each category has videos of varying road, traffic and weather conditions.

We show the accuracy of self-positioning before and after temporal smoothing in Table \ref{table:accuracy}. Table \ref{table:RFConfMat} shows the confusion matrix for the initial estimation of self-positioning and the final estimate after temporal smoothing. We can see that there is an improvement of $5.04\%$ over all $8$ categories, which indicates that temporal smoothing is able to remove noisy results persisting only for a few frames. As we increase the value of $p$ in equation \ref{eq:genFormulatn}, the algorithm averages over larger set of frames but it increases the memory and makes the algorithm computationally inefficient. Thus we have set the value of $p$ to $15$ while using temporal smoothing, otherwise we make it $0$. Temporal smoothing improves the initial estimation in all but one category -$[4,1]$. The decrease may occur when the algorithm wrongly predicts for more than $p/2$ frames often. Also, due to slowly increasing exponential kernel, first $p/2$ frames may be wrongly predicted when the category changes. Low prediction accuracy is observed in the category -$[4,1]$. It should be accounted that there are 3 lanes on one of the sides of the host-lane. Since its extremely thin (about 4 pixels), its reliable detection becomes difficult. In addition to this, the road conditions in that category are relatively worse and there is moderate traffic. Interestingly, the category $[6,4]$ achieves $96.36\%$ accuracy in spite of having most number of lanes. We would like to point out the fact that this category has the best weather and road conditions. Thus our approach can achieve high accuracy even in the presence of more number of lanes provided the weather, road and traffic conditions permit. This is an eight-class classification problem an it has variable weather and road conditions. Another important fact is that the training data itself may be noisy. We have divided the video clips such that each clip has a single label. Therefore, one of the training video clips in category - $[5,3]$ (say) may have a faint/invisible second lane for a few frames. Results using our approach are shown in Fig. \ref{fig:rightWrongDetections}. The proposed approach is able to handle partial vehicle occlusions, varying road and weather conditions. However, it may fail when there are multiple vehicles totally blocking the view or if other conditions are worse.

Bounding box selection technique reduces the generated proposals for vehicles by a factor of 6 and in turn this reduces number of false detections. This technique also allows use of other faster but less accurate methods. We would like to point out that though vehicle detection is an important component in our system, our final goal remains to achieve accurate self-positioning in all possible conditions. We do not yet have annotated database of vehicles for such a scenario and thus we do not list comprehensive results for vehicle detection.

\section{Conclusion and Future Scope}

A novel problem of self-positioning is introduced and an integrated approach is proposed which performs self-positioning with the aid of vehicle detection in a closed-loop system. A high-level Bayesian formulation is developed to allow easy modifications in the future should anyone try to introduce additional factors such as scene or viewpoint information. Our approach enables a system to perform dynamic traffic routing on lane-level due to its knowledge of vehicle positions and lane-structure. This promises larger reduction in congestion cost and travel delays. We also develop a bounding box selection criteria which can be applied to exhaustive set of boxes or to the boxes obtained from other box-proposal methods. Testing this framework on real-world videos yielded acceptable results.

The system presently uses just video data to make predictions. In future, we can use GPS, accelerometer data and other information obtained from vehicle dynamics. It is also possible to employ this system on cloud and integrate the results from all the vehicles to understand the underlying true structure of the road. Additional information such as position of lanes traveling in opposite direction may be included so that detection of those vehicles could be avoided. We also aim to generate an annotated database of vehicles in order to perform vehicle density estimation along with self-positioning and evaluate the same.

{\small
\bibliographystyle{ieee}
\bibliography{IEEEexample}
}

\end{document}